\documentclass[9pt,twocolumn,twoside]{osajnl}
\usepackage{times}
\usepackage{amssymb}
\usepackage{multirow}
\usepackage{mathtools}
\usepackage{subfloat}
\usepackage{bbm}
\usepackage{algorithm}
\usepackage{algpseudocode}
\usepackage{pbox}
\usepackage{placeins}
\journal{ol} 
\setboolean{shortarticle}{false}

\usepackage{times}
\usepackage{graphicx}
\usepackage{units}
\usepackage{url}

\usepackage{amsmath}
\renewcommand{\vec}{\boldsymbol}

\usepackage[printonlyused]{acronym}

\acrodef{RL}{reinforcement learning}
\acrodef{TO}{trajectory optimization}
\acrodef{PPO}{Proximal Policy Optimization}
\acrodef{MLP}{multilayer perceptron}
\acrodef{MPC}{model predictive control}
\acrodef{SOA}{state-of-the-art}
\acrodef{TAMOLS}{terrain-aware motion generation for legged systems}
\acrodef{DTC}{Deep Tracking Control}
\acrodef{IK}{inverse kinematics}
\acrodef{WBC}{whole-body control}

\ifx\isReview\undefined
    \newcommand\highlight[1]{#1}
\else
    \usepackage{xcolor}
    \newcommand\highlight[1]{\textcolor{red}{#1}}
\fi

\title{DTC: Deep Tracking Control} 

\author{
Fabian~Jenelten,$^{1\ast}$ 
Junzhe~He,$^{1}$ 
Farbod~Farshidian,$^{2}$ 
and~Marco~Hutter$^{1}$ 
\\
\normalsize{$^{1}$Robotic Systems Lab, ETH Zurich, 8092 Zurich, Switzerland.} \\
\normalsize{$^{2}$Currently at Boston Dynamics AI Institute, 145 Broadway, Cambridge MA, USA} \\
\normalsize{$^\ast$Corresponding author: fabian.jenelten@ethz.ch}
}

\dates{This is the accepted version of Science Robotics Vol. 9, Issue 86, eadh5401 (2024) \\ DOI: 0.1126/scirobotics.adh5401}

\begin{abstract}
Legged locomotion is a complex control problem that requires both accuracy and robustness to cope with real-world challenges. Legged systems have traditionally been controlled using trajectory optimization with inverse dynamics. Such hierarchical model-based methods are appealing due to intuitive cost function tuning, accurate planning, generalization, and most importantly, the insightful understanding gained from more than one decade of extensive research. However, model mismatch and violation of assumptions are common sources of faulty operation. Simulation-based reinforcement learning, on the other hand, results in locomotion policies with unprecedented robustness and recovery skills.
Yet, all learning algorithms struggle with sparse rewards emerging from environments where valid footholds are rare, such as gaps or stepping stones. In this work, we propose a hybrid control architecture that combines the advantages of both worlds to simultaneously achieve greater robustness, foot-placement accuracy, and terrain generalization. Our approach utilizes a model-based planner to roll out a reference motion during training. A deep neural network policy is trained in simulation, aiming to track the optimized footholds. We evaluate the accuracy of our locomotion pipeline on sparse terrains, where pure data-driven methods are prone to fail. Furthermore, we demonstrate superior robustness in the presence of slippery or deformable ground when compared to model-based counterparts. Finally, we show that our proposed tracking controller generalizes across different trajectory optimization methods not seen during training. In conclusion, our work unites the predictive capabilities and optimality guarantees of online planning with the inherent robustness attributed to offline learning.
\end{abstract}

\begin{document} 
\maketitle

\section*{Introduction}
\begin{figure*}
\centering
\includegraphics[width=7.3in]{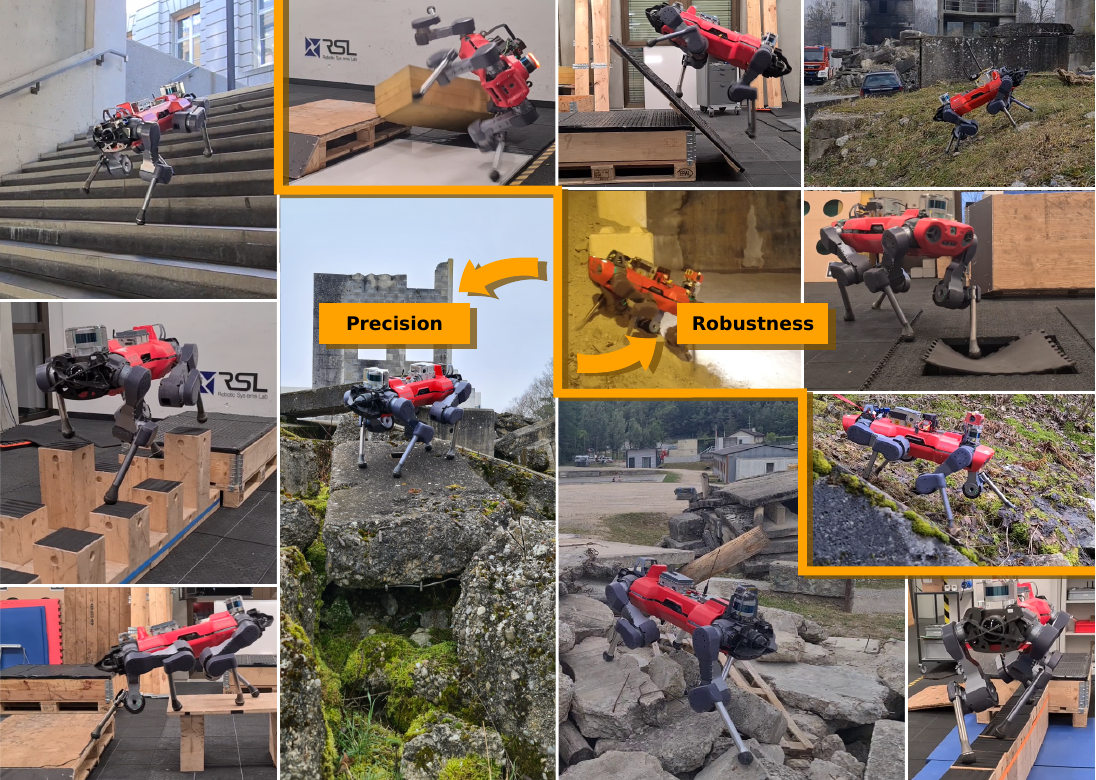}
\caption{\textbf{\highlight{Robust and precise locomotion in various indoor and outdoor environments.}} \highlight{The marriage of model-free and model-based control} allows legged robots to be deployed in environments where steppable contact surfaces are sparse (bottom left) and environmental uncertainties are high (top right).}
\label{fig:intro:intro}
\end{figure*}
\Ac{TO} is a commonly deployed instance of optimal control for designing motions of legged systems and has a long history of successful applications in rough environments since the early 2010s~\cite{Kolter2008,Kalakrishnan2010}. These methods require a model of the robot's kinematics and dynamics during runtime, along with a parametrization of the terrain. Until recently, most approaches have used simple models such as single rigid body~\cite{Winkler2018} or inverted pendulum dynamics~\cite{Mastalli2020,Jenelten2020}, or have ignored the dynamic effects altogether~\cite{Fankhauser2018}. Research has shifted towards more complex formulations, including centroidal~\cite{Ruben2022} or full-body dynamics~\cite{Mastalli2022}. The resulting trajectories are tracked by a \ac{WBC} module, which operates at the control frequency and utilizes full-body dynamics~\cite{Bellicoso2}. Despite the diversity and agility of the resulting motions, there remains a \highlight{considerable} gap between simulation and reality due to unrealistic assumptions. Most problematic assumptions include perfect state estimation,  occlusion-free vision, known contact states, zero foot-slip, and perfect realization of the planned motions. Sophisticated hand-engineered state machines are required to detect and respond to various special cases not accounted for in the modeling process. Nevertheless, highly dynamic jumping maneuvers performed by Boston Dynamics' bipedal robot Atlas demonstrate the potential power of \ac{TO}.

\Ac{RL} has emerged as a powerful tool in recent years for synthesizing robust legged locomotion. Unlike model-based control, \ac{RL} does not rely on explicit models. Instead, behaviors are learned, most often in simulation, through random interactions of agents with the environment. The result is a closed-loop control policy, typically represented by a deep neural network, that maps raw observations to actions. Handcrafted state-machines become obsolete because all relevant corner cases are eventually visited during training. End-to-end policies, trained from user commands to joint target positions, have been deployed successfully on quadrupedal robots such as ANYmal~\cite{Hwangbo2019,Rudin2021}. More advanced teacher-student structures have \highlight{substantially} improved the robustness, enabling legged robots to overcome obstacles through touch~\cite{Lee2020} and perception~\cite{Miki2022}. \highlight{Although} locomotion \highlight{across} gaps and stepping stones is theoretically possible, good exploration strategies are required to learn from the emerging sparse reward signals. So far, these terrains could only be handled by specialized policies, which intentionally overfit to one particular scenario~\cite{Rudin2022} or a selection of similar terrain types~\cite{allsteps,Duan2022,Yu2021,Agarwal2022}.
Despite promising results, distilling a unifying locomotion policy may be difficult and has only been shown with limited success~\cite{caluwaerts2023barkour}.

Some of the shortcomings that appear in \ac{RL} can be mitigated using optimization-based methods. While the problem of sparse gradients still exists, two important advantages can be exploited: First, cost-function and constraint gradients can be computed with a small number of samples. Second, poor local optima can be avoided by pre-computing footholds~\cite{Jenelten2020,Mastalli2022}, pre-segmenting the terrain into steppable areas~\cite{Griffin2019,Ruben2022}, or by smoothing out the entire gradient landscape~\cite{Jenelten2022}. Another advantage of \ac{TO} is its ability to plan actions ahead and predict future interactions with the environment. If model assumptions are generic enough, this allows for great generalization across diverse terrain geometries~\cite{Jenelten2022,Ruben2022}.

The sparse gradient problem has been addressed extensively in the learning community. A notable line of research has focused on learning a specific task while imitating expert behavior. The expert provides a direct demonstration for solving the task~\cite{Brakel2021,Bogdanovic2022}, or is used to impose a style while discovering the task~\cite{Peng2018,Peng2021,Bohez2022}. These approaches require collecting expert data, commonly done offline, either through re-targeted motion capture data~\cite{Peng2018,Peng2021,Bohez2022} or a \ac{TO} technique~\cite{Brakel2021,Bogdanovic2022}. The reward function can now be formulated to be dense, meaning that agents can collect non-trivial rewards even if they do not initially solve the task. Nonetheless, the goal is not to preserve the expert's accuracy but rather to lower the sample and reward complexity by leveraging existing knowledge.

To further decrease the gap between the expert and the policy performance, we speculate that the latter should have insight into the expert's intentions. This requires online generation of expert data, which can be conveniently achieved using any model-based controller. Unfortunately, rolling out trajectories is often orders of magnitude more expensive than a complete learning iteration. To circumvent this problem, one possible alternative is to approximate the expert with a generative model, \highlight{for instance}, by sampling footholds from a uniform distribution~\cite{allsteps,Duan2022}, or from a neural network~\cite{Tsounis2020,deeploco,Yu2021}. However, for the former group, it might be challenging to capture the distribution of an actual model-based controller, while the latter group still does not solve the exploration problem itself.

In this work, we propose to guide exploration through the solution of \ac{TO}. As such data will be available both on- and offline, we refer to it as ``reference'' and not expert motion. We utilize a hierarchical structure introduced in deep loco~\cite{deeploco}, where a high-level planner proposes footholds at a lower rate, and a low-level controller follows the footholds at a higher rate. Instead of using a neural network to generate the foothold plan, we leverage \ac{TO}. Moreover, we do not only use the target footholds as an indicator for a rough high-level direction but as a demonstration of optimal foot placement. 

The idea of combining model-based and model-free approaches is not new in the literature. For instance, supervised~\cite{towr_learning} and unsupervised~\cite{Surovik2021, Melon2021} learning has been used to warm-start nonlinear solvers. \ac{RL} has been used to imitate~\cite{Bogdanovic2022,Brakel2021} or correct~\cite{Gangapurwala2022} motions obtained by solving \ac{TO} problems. Conversely, model-based methods have been used to check the feasibility of learned high-level commands~\cite{Tsounis2020} or to track learned acceleration profiles~\cite{glide}. Compared to~\cite{Gangapurwala2022}, we do not learn corrective joint torques around an existing \ac{WBC}, but instead, learn the mapping from reference signals to joint positions in an end-to-end fashion. 

To generate the reference data, we rely on an efficient \ac{TO} method called \ac{TAMOLS}~\cite{Jenelten2022}. It optimizes over footholds and base pose simultaneously, thereby enabling the robot to operate at its kinematic limits. We let the policy observe only a small subset of the solution, namely planar footholds, desired joint positions, and the contact schedule. We found that these observations are more robust under the common pitfalls of model-based control, while still providing enough information to solve the locomotion task. In addition, we limit computational costs arising from solving the optimization problems by utilizing a variable update rate. During deployment, the optimizer runs at the fastest possible rate to account for model uncertainties and external disturbances.

Our approach incorporates elements introduced in~\cite{Rudin2022}, such as time-based rewards and position-based goal tracking. However, we reward desired foothold positions at planned touch-down instead of rewarding a desired base pose at an arbitrarily chosen time. Finally, we use an asymmetric actor-critic structure similar to~\cite{Brakel2021}, where we provide privileged ground truth information to the value function and noisified measurements to the network policy. 

We trained more than $4000$ robots in parallel for two weeks on challenging ground covering a surface area of more than $\unit[76000]{m^2}$. Throughout the entire training process, we generated and learned from about $23$ years of optimized trajectories. The combination of offline training and online re-planing results in \highlight{accurate, agile, and robust locomotion}. As showcased in Fig.~\ref{fig:intro:intro} and movie~1, with our hybrid control pipeline, ANYmal~\cite{Hutter2016} can skillfully traverse parkours with high precision, and confidently overcome uncertain environments with high robustness. Without the need for any post-training, the tracking policy can be deployed zero-shot with different \ac{TO} methods at different update rates. \highlight{Moreover, movie~2 demonstrates successful deployment in search-and-rescue scenarios, which demand both accurate foot placement and robust recovery skills.} 
The contributions of our work are therefore twofold: Firstly, we enable the deployment of model-based planners in rough and uncertain real-world environments. Secondly, we create a single unifying locomotion policy that generalizes beyond the limitations imposed by state-of-the-art \ac{RL} methods.

\section*{Results}
In order to evaluate the effectiveness of our proposed pipeline, hereby referred to as \ac{DTC}, we compared it with four different approaches: two model-based controllers, \ac{TAMOLS}~\cite{Jenelten2022} and a nonlinear \ac{MPC} method presented in~\cite{Ruben2022}, and two data-driven methods, as introduced in~\cite{Miki2022} and~\cite{Rudin2021}. We refer to those as baseline-to-1 (\ac{TAMOLS}), baseline-to-2 (\ac{MPC}), baseline-rl-1 (teacher/student policy), and baseline-rl-2 (\ac{RL} policy),  respectively. These baselines mark the state-of-the-art in \ac{MPC} and \ac{RL} prior to this work and they have been tested and deployed under various conditions. If not noted differently, all experiments were conducted in the real world.

\subsection*{Evaluation of Robustness}
We conducted three experiments to evaluate the robustness of our hybrid control pipeline. The intent is to demonstrate survival skills on slippery ground, and recovery reflexes when visual data is not consistent with proprioception or is absent altogether. We \highlight{rebuilt} harsh environments that are likely to be encountered on sites of natural disasters, where debris might further break down when stepped onto, and construction sites, where oil patches create slippery surfaces.

\begin{figure*}
\centering
\includegraphics[width=7.3in]{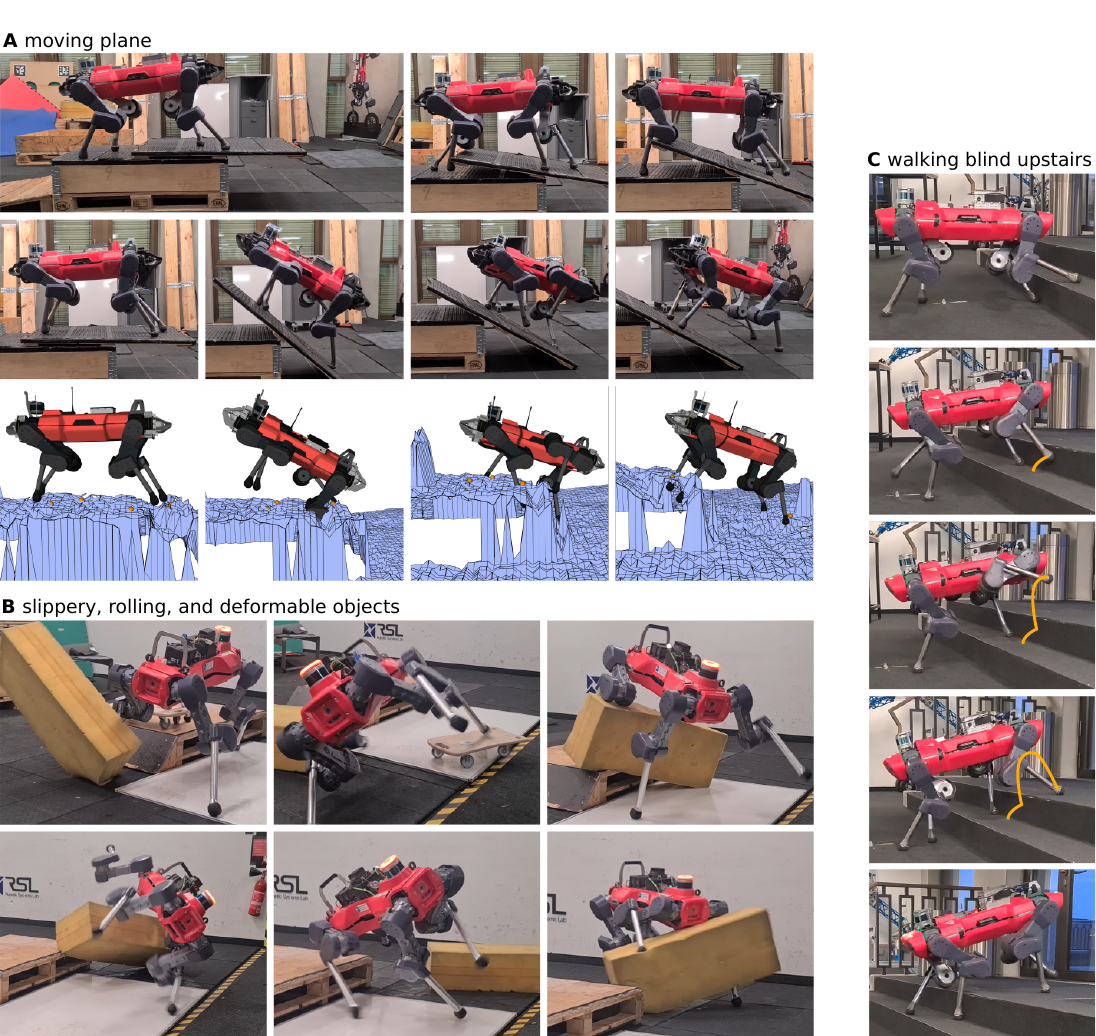}
\caption{\textbf{Evaluation of robustness.} \textbf{(A)} ANYmal walks along a loose cover plate that eventually pitches forward (left to right, top to bottom). The third row shows ANYmal's perception of the surroundings during the transition and recovery phase. \textbf{(B)} The snapshots are taken at critical time instances when walking on slippery ground, just before complete recovery. \textbf{(C)} ANYmal climbs upstairs with disabled perception (top to bottom). The collision of the right-front end-effector with the stair tread triggers a swing reflex, visualized in orange.}
\label{fig:robustness}
\end{figure*}

In the first experiment, we placed a rectangular cover plate with an area of $\unit[0.78 \times 1.19]{m^2}$ on top of a box with the same length and width, and height $\unit[0.37]{m}$ (Fig.~\ref{fig:robustness} A). The cover plate was shifted to the front, half of the box's length. ANYmal was then steered over the cover plate, which pitched down as soon as its center of mass passed beyond the edge of the box. Facing only forward and backward, the plate's movement was not detected through the depth cameras, and could only be perceived through proprioceptive sensors. Despite the error between map and odometry reaching up to $\unit[0.4]{m}$, the robot managed to successfully balance itself. This experiment was repeated three times with consistent outcomes.

In our second experiment (Fig.~\ref{fig:robustness} B) we created an obstacle parkour with challenging physical properties. A large wooden box with a slopped front face was placed next to a wet and slippery whiteboard. We increased the difficulty by placing a soft foam box in front, and a rolling transport cart on top of the wooden box. The robot was commanded to walk over the objects with random reference velocities for approximately $45$ seconds, after which the objects were redistributed to their original locations to account for any potential displacement. This experiment was repeated five times. Despite not being trained on movable or deforming obstacles, the robot demonstrated its recovery skills in all five trials without any falls. 

The tracking policy was trained with perceptive feedback, meaning that the policy and the motion planner had partial or complete insight into the local geometrical landscape. Nevertheless, the locomotion policy was still capable of overcoming many obstacles completely blind. To simulate a scenario with damaged depth sensors, we let ANYmal blindly walk over a stair with two treads, each $\unit[0.18]{m}$ high and $\unit[0.29]{m}$ wide (Fig.~\ref{fig:robustness} C). The experiment was repeated three times up and down, with an increasing heading velocity selected from $\unit[\{\pm0.5, \pm0.75, \pm1.0\}]{m/s}$. In some cases, a stair tread was higher than the swing motion of a foot. Thanks to a learned swing reflex, the stair set could be successfully cleared in all trials. We note that the same stair set was passed by a blindfolded version of baseline-rl-1~\cite{Miki2022}, which was trained in a complex teacher/student environment. In contrast, our method relies on an asymmetric actor/critics structure, achieving a similar level of robustness. Accompanying video clips can be found in the supplementary movie S1.

\subsection*{Evaluation of Accuracy}
\begin{figure*}
\centering
\includegraphics[width=7.3in]{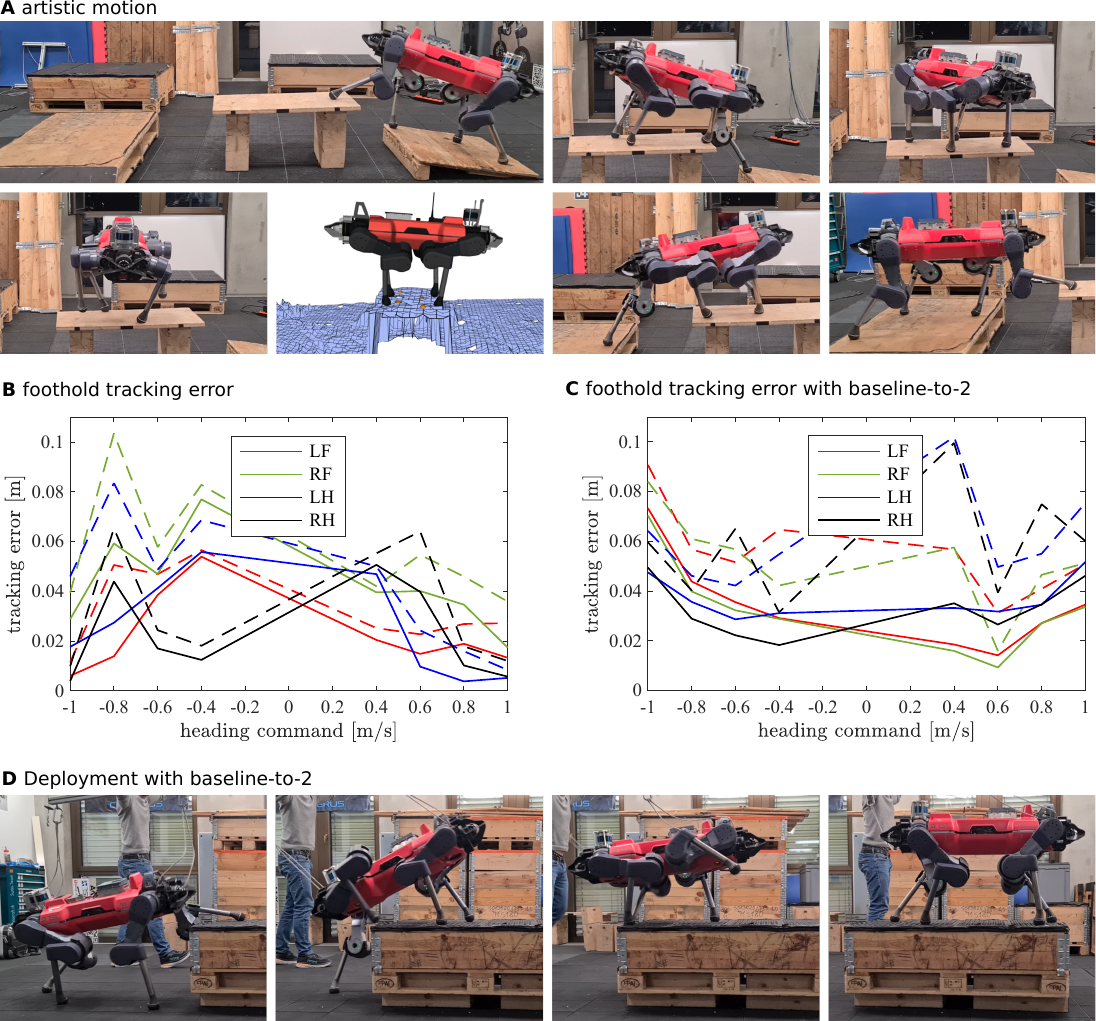}
\caption{\textbf{Evaluation of tracking performance.} \textbf{(A)} ANYmal climbs up a narrow table, turns, and descends back down to a box. The second image in the second row shows the robot's perception of the environment.
\textbf{(B)} Euclidean norm of the planar foothold error, averaged over $\unit[20]{s}$ of operation using a constant heading velocity. The solid/dashed curves represent the average/maximum tracking errors. \textbf{(C)} Same representation as in (B), but the data was collected with baseline-to-2.
\textbf{(D)} \ac{DTC} deployed with baseline-to-2, enabling ANYMal to climb up a box of $\unit[0.48]{m}$.}
\label{fig:accuracy}
\end{figure*}

We demonstrate the precision of foothold tracking by devising a complex motion that requires the robot to perform a turn-in-place maneuver on a small surface area of $\unit[0.94\times 0.44]{m^2}$. The robot was commanded to walk up a slope onto a narrow table, then to execute a complete $\unit[360]{deg}$ turn, and finally to descend onto a pallet. Some snapshots of the experiment are provided in Fig.~\ref{fig:accuracy} A, \highlight{whereas} the full video is contained in movie S2.

To evaluate the quality of the foothold tracking, we collected data while ANYmal walked on flat ground. Each experiment lasted for approximately $\unit[20]{s}$ and was repeated with eight different heading velocities selected from $\unit[\{\pm1.0, \pm0.8, \pm 0.6, \pm0.4\}]{m/s}$. We measured the tracking error as the smallest horizontal distance between a foot and its associated foothold during a stance phase. As shown in Fig.~\ref{fig:accuracy} B, the footholds could be tracked with very high precision of $\unit[2.3]{cm}$ and standard deviation $\unit[0.48]{cm}$ when averaged over the broad spectrum of heading velocity commands.

\subsection*{Deployment with \ac{MPC}}
The maximum height that \ac{DTC} in combination with \ac{TAMOLS} can reliably overcome is about $\unit[0.40]{m}$. The policy might hesitate to climb up taller objects due to the risk of potential knee joint collisions with the environment. This limitation is inherent to the chosen \ac{TO} method, which only considers simplified kinematic constraints. We, therefore, deployed \ac{DTC} with the planner of baseline-to-2, which takes into account the full kinematics of the system. To allow for zero-shot generalization, we implemented the same trotting gait as experienced during training. With this enhanced setup, ANYMal could climb up a box of height of $\unit[0.48]{m}$. This is $\unit[50]{\%}$ higher than what baseline-rl-1 \highlight{could} climb up, and $\unit[380]{\%}$ more than what was reported for baseline-rl-2. The box climbing experiment was successfully repeated five times. The results are shown in movie S2, and for one selected trial in Fig.~\ref{fig:accuracy} D. Furthermore, we measured the tracking error on flat ground. Despite the wider stance configuration of baseline-to-2, the error was found to be only $\unit[0.03]{m}$ on average (Fig.~\ref{fig:accuracy} C).

The above two results seem to be surprising at first glance but are easy to explain when considering the observation space and the training environment. \highlight{Although} the base-pose trajectory is considerably more detailed for baseline-to-2 due to frequency-loop shaping and increased system complexity, the foothold patterns are nevertheless quite similar. Thus, good generalization is facilitated by the specific choice of observations, which hides the optimized base pose from the policy. 
Some terrains \highlight{within the training environment} can be seen as a combination of gaps and boxes, where each box is surrounded by a gap. During training, \ac{TAMOLS} placed the footholds sufficiently far away from the box to avoid stepping into the gap. This allowed the policy to learn climbing maneuvers without knee joint collisions. Baseline-to-2, being aware of the spatial coordinates of the knees, naturally produces a similar foothold pattern, even in the absence of the gap.

\subsection*{Benchmark Against Model-Based Control}
\begin{figure*}
\centering
\includegraphics[width=7.3in]{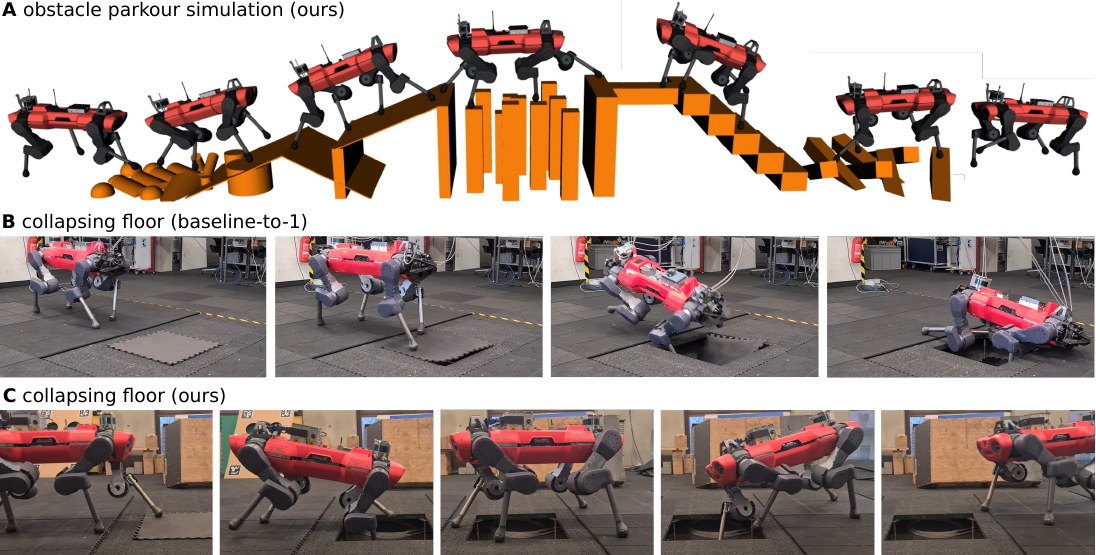}
\caption{\textbf{\highlight{Benchmarking} against model-based control.} \textbf{(A)} \ac{DTC} successfully traverses an obstacle parkour (left to right) in simulation with a heading velocity of $\unit[1]{m/s}$. \textbf{(B)} Baseline-to-1 falls after stepping into a gap hidden from the perception (left to right). \textbf{(C)} ANYmal successfully overcomes a trapped floor using our hybrid control architecture (left to right).}
\label{fig:results:baseline_mpc}
\end{figure*}

\ac{TO} was proven to be effective in solving complex locomotion tasks in simulation, such as the parkour shown in Fig.~\ref{fig:results:baseline_mpc} A. This parkour has been successfully traversed by ANYmal using baseline-to-1 and baseline-to-2, while it was found to be non-traversable for baseline-rl-1 and baseline-rl-2~\cite{Ruben2022}. With our proposed approach, ANYmal could cross the same obstacle parkour in simulation back and forth at a speed of $\unit[1]{m/s}$, which \highlight{was} $\unit[20]{\%}$ faster than baseline-to-1. The corresponding video clip can be found in movie S3.

Model-based controllers react sensitively to violation of model assumptions, which hinders applications in real-world scenarios, where, for instance, uncertainties in friction coefficients, contact states, and visual perception may be large. This issue is exemplified in Fig.~\ref{fig:results:baseline_mpc} B, where baseline-to-1 was used to guide ANYmal over a flat floor with an invisible gap. When the right front foot stepped onto the trap, the planned and executed motions deviated from each other. This triggered a sequence of heuristic recovery strategies. For large mismatches, however, such scripted reflexes \highlight{were} not effective, and \highlight{resulted} in failure. \ac{DTC} uses the same high-level planner but incorporates learned recovery and reflex skills. \highlight{This allowed} the robot to successfully navigate through the trap. The robustness is rooted in the ability to ignore both perception and reference motion while relying only on proprioception. Such behavior \highlight{was} learned in simulation by experiencing simulated map drift. The experiment was repeated five times with baseline-to-1, five times with baseline-to-2, and five times with our method, consistently leading to similar results. The video clips corresponding to the above experiments can be found in movie S3. The movie is further enriched with a comparison of baseline-to-2 against \ac{DTC} on soft materials, which impose very similar challenges.

\subsection*{Benchmark Against RL Control}
\begin{figure*}
\centering
\includegraphics[width=5in]{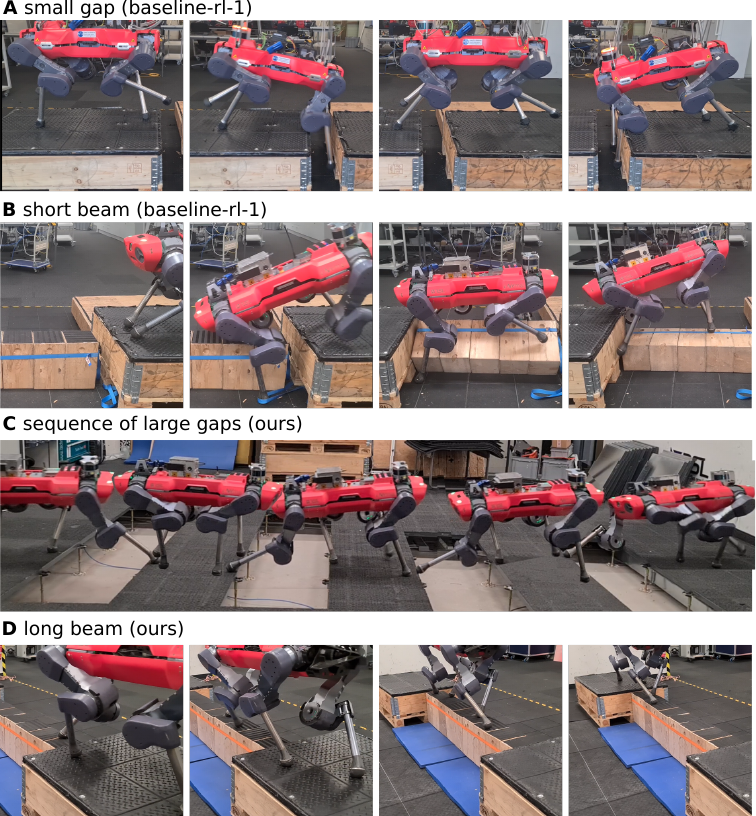}
\caption{\textbf{\highlight{Benchmarking} against reinforcement learning.} \textbf{(A)} Baseline-rl-1 attempts to cross a small gap. ANYmal initially manages to recover from miss-stepping with its front legs but subsequently gets stuck as its hind legs fall inside the gap. \textbf{(B)} Using baseline-rl-1, the robot stumbles along a narrow beam. \textbf{(C)} With \ac{DTC}, the robot can pass four consecutive large gaps (left to right) without getting stuck or falling. \textbf{(D)} ANYmal is crossing a long beam using the proposed control framework.}
\label{fig:baseline_rl}
\end{figure*}

\highlight{Although} \ac{RL} policies are known for their robustness, they may struggle in environments with limited interaction points. We demonstrate typical failure cases in two experiments utilizing baseline-rl-1. In the first experiment (Fig.~\ref{fig:baseline_rl} A), ANYmal was tasked to cross a small gap of $\unit[0.1]{m}$ with a reference heading velocity of $\unit[0.2]{m/s}$. The model-free controller did not avoid the gap, and thus could not reach the other side of the platform. In the second experiment, we connected two elevated boxes with a $\unit[1.0]{m}$-long beam of height $\unit[0.2]{m}$ (Fig.~\ref{fig:baseline_rl} B). The robot was commanded to walk from the left to the right box but failed to make use of the beam.

In comparison, our hybrid policy achieved a $\unit[100]{\%}$ success rate for the same gap size over ten repetitions.
To further demonstrate the locomotion skills of \ac{DTC}, we made the experiments more challenging. We replaced the small gap with four larger gaps, each $\unit[0.6]{m}$ wide and evenly distributed along the path (Fig.~\ref{fig:baseline_rl} C). Similarly, we increased the length of the beam to a total of $\unit[1.8]{m}$  (Fig.~\ref{fig:baseline_rl} D). Despite the increased difficulty, our approach maintained a $\unit[100]{\%}$ success rate across four repetitions of each experiment. Video clips of those experiments can be found in movie S4.

By using a specialized policy, ANYmal crossed a $\unit[0.6]{m}$ wide gap within a pre-mapped environment~\cite{Rudin2022}. Most notably, our locomotion controller, not being specialized nor fine-tuned for this terrain type, crossed a sequence of four gaps with the same width, \highlight{whilst}, relying on online generated maps only.

\begin{figure*}
\centering
\includegraphics[width=3.5in]{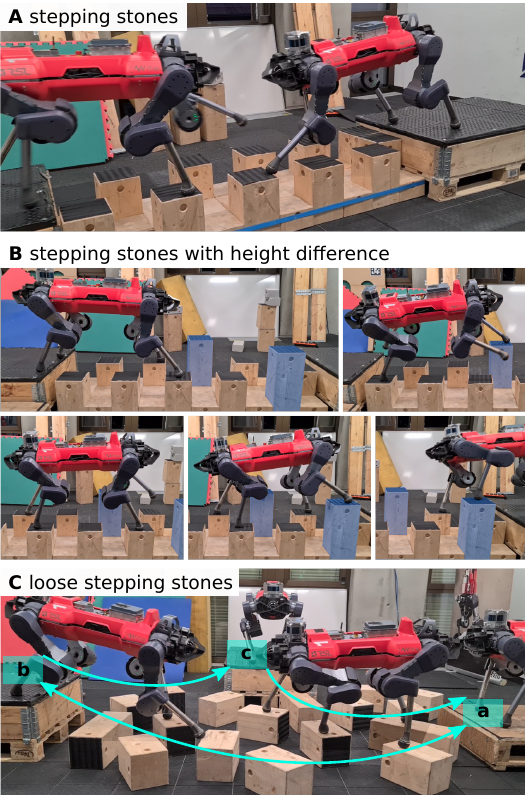}
\caption{\textbf{Evaluation of the locomotion performance on stepping stones.} \textbf{(A)} ANYmal reliably crosses a field of flat stepping stones (left to right). \textbf{(B)} The robot crosses stepping stones of varying heights (left to right). The two tall blocks are highlighted in blue. \textbf{(C)} ANYmal navigates through a field of loosely connected stepping stones, following the path $a\to b\to a \to b \to c \to a$.}
\label{fig:stepping_stones}
\end{figure*}

The limitations of baseline-rl-1 were previously demonstrated~\cite{Ruben2022} on the obstacle parkour of Fig.~\ref{fig:results:baseline_mpc} A, showing its inability to cross the stepping stones. We showcase the generality of our proposed control framework by conducting three experiments on stepping stones in the real world, each with an increased level of difficulty. The first experiment (Fig.~\ref{fig:stepping_stones} A) required ANYmal traversing a field of equally sized stepping stones, providing a contact surface of $\unit[0.2 \times 0.2]{m^2}$ each. The robot passed the $\unit[2.0]{m}$ long field $10$ times. Despite the varying heading velocity commands, the robot accurately hit the correct stepping stones as indicated by the solution of the \ac{TO}. For the second experiment (Fig.~\ref{fig:stepping_stones} B), we increased the height of two randomly selected stones. The parkour was successfully crossed four out of four times. In the final experiment (Fig.~\ref{fig:stepping_stones} C), we \highlight{distributed} three elevated platforms $a$, $b$, and $c$, connected by loose wooden blocks of sizes $\unit[0.31\times0.2\times0.2]{m^3}$ and $\unit[0.51\times0.2\times0.2]{m^3}$. This environment posed \highlight{considerable} challenges as the blocks may move and flip over when stepped on. Following the path $a\to b\to a \to b \to c \to a$, the robot missed only one stepping stone, which, however, did not lead to failure. Video clips of the stepping stones experiments are provided in movie S5.

\subsection*{Simulation-Based Ablation Study}
\begin{figure*}
\centering
\includegraphics[width=7.3in]{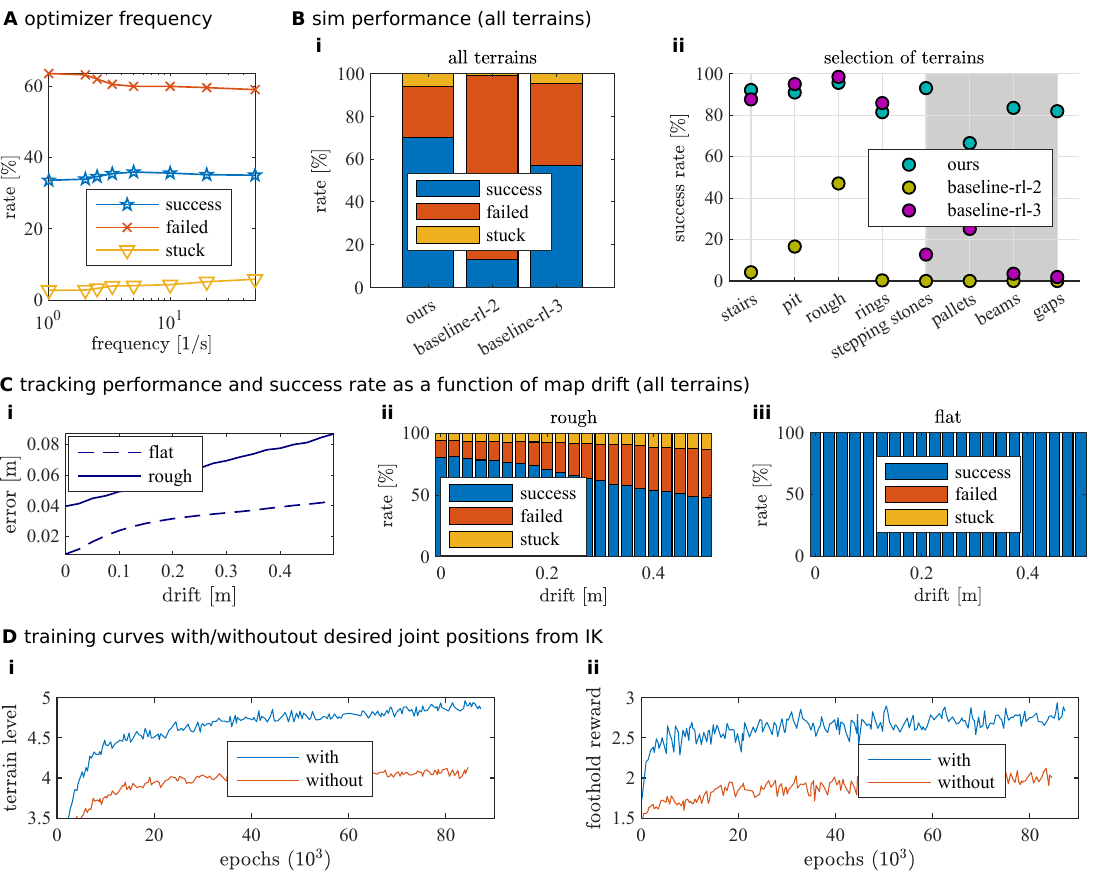}
\caption{\textbf{Simulation results and ablation studies}. \textbf{(A)} Success and failure rates of \ac{DTC}, recorded for different update rates of the optimizer. The upper limit of $\unit[50]{Hz}$ is imposed by the policy frequency. \textbf{(B)} Comparison against baseline policies. \textbf{(i)} Evaluation on all $120$ terrains. \textbf{(ii)} Evaluation on terrains where valid footholds are dense (white background) and sparse (gray background). \textbf{(C)} Influence of elevation map drift on the locomotion performance, quantified by tracking error \textbf{(i)}, success rate on rough \textbf{(ii)}, and on flat ground \textbf{(ii)}. \textbf{(D)} Average terrain level \textbf{(i)} and average foothold reward \textbf{(ii)} scored during training.}
\label{fig:sim}
\end{figure*}

During training, we \highlight{computed} a solution to the \ac{TO} problem after variable time intervals, but mainly after each foot touch-down. \highlight{Although} such a throttled rate greatly \highlight{reduced} computational costs, it also leads to poor reactive behavior in the presence of quickly changing external disturbances, dynamic obstacles, or map occlusion. Moreover, the optimizer was updated using privileged observations, whereas, in reality, the optimizer is subject to elevation map drift, wrongly estimated friction coefficients, and unpredicted external forces. To compensate for such modeling errors, we deploy the optimizer in \ac{MPC}-fashion. We investigated the locomotion performance as a function of the optimizer update rate. Using the experimental setup outlined in \highlight{the supplementary methods (section ``Experimental Setup for Evaluation of Optimizer Rate'')}, we collected a total of six days of data in simulation. A robot was deemed ``successful'' if it could walk from the center to the border of its assigned terrain patch, ``failed'' if its torso made contact with the environment within its patch, and ``stuck'' otherwise. We report success and failure rates in Fig.~\ref{fig:sim} A. Accordingly, when increasing the update rate from $\unit[1]{Hz}$ to $\unit[50]{Hz}$, the failure rate dropped by $\unit[7.11]{\%}$ \highlight{whereas} the success rate increased by $\unit[4.25]{\%}$. 

In the second set of experiments, we compared our approach against baseline-rl-2 as well as against the same policy trained within our training environment. We refer to the emerging policy as baseline-rl-3. More details regarding the experimental setup can be found in \highlight{the supplementary methods (section ``Experimental Setup for Performance Evaluation'')}. As depicted in Fig.~\ref{fig:sim} B i, our approach \highlight{exhibited} a substantially higher success rate than baseline-rl-2. By learning on the same terrains, baseline-rl-3 \highlight{could} catch up but still \highlight{did} not match our performance \highlight{in terms of success rate}. The difference mainly originates from the fact that the retrained baseline still failed to solve sparse-structured terrains. To highlight this observation, we evaluated the performance on four terrain types with sparse (``stepping stones'', ``beams'', ``gaps'', and ``pallets''), and on four types with dense stepping locations (``stairs'', ``pit'', ``rough slope'', and ``rings'').
On all considered terrain types, our approach \highlight{outperformed} baseline-rl-2 by a huge margin (Fig.~\ref{fig:sim} B ii), thereby demonstrating that learned locomotion generally does not extrapolate well to unseen scenarios. We \highlight{performed} equally well as baseline-rl-3 on dense terrains, but \highlight{scored notably} higher on sparse-structured terrains. This result suggests that the proposed approach itself \highlight{was} effective and that favorable locomotion skills \highlight{were} not encouraged by the specific training environment.

In an additional experiment, we investigated the \highlight{effects} of erroneous predictions of the high-level planner on locomotion performance. We did so by adding a drift value to the elevation map, sampled uniformly from the interval $\in\unit[(0,0.5)]{m}$. 
Contrary to training, the motion was optimized over the perturbed height map. Other changes to the experimental setup are described in \highlight{the supplementary methods (section ``Experimental Setup for Performance Evaluation under Drift'')}. As visualized in Fig.~\ref{fig:sim} C, we collected tracking error, success, and failure rates with simulated drift on flat and rough ground. The tracking error \highlight{grew} mostly linearly with the drift value (Fig.~\ref{fig:sim} C i). On rough terrains, the success rate \highlight{remained} constant for drift values smaller than $\unit[0.1]{m}$, and decreased linearly for larger values (Fig.~\ref{fig:sim} C ii). On the other hand, success and failure rates \highlight{were not affected} by drift on flat ground (Fig.~\ref{fig:sim} C iii).

We found that providing joint positions computed for the upcoming touch-down event greatly \highlight{improved} convergence time and foothold tracking performance. This signal encodes the foothold location in joint space, thus, providing a useful hint for foothold tracking. It also simplifies the learning process, as the network is no \highlight{longer} required to implicitly learn the \ac{IK}. Evidence for our claims is given in Fig.~\ref{fig:sim} D, showing two relevant learning curves. Tracking accuracy is represented by the foothold rewards, whereas technical skills are quantified using the average terrain level~\cite{Rudin2021}. Both scores are substantially higher if the footholds \highlight{could} be observed in both task and joint space.

\section*{Discussion}
This work demonstrates the potential of a hybrid locomotion pipeline that combines accurate foot placement and dynamic agility of state-of-the-art \ac{TO} with the inherent robustness and reflex behaviors of novel \ac{RL} control strategies. Our approach enables legged robots to overcome complex environments that either method alone would struggle with. As such terrains are commonly found in construction sites, mines, and collapsed buildings, our work could help advance the deployment of autonomous legged machines in the fields of construction, maintenance, and search-and-rescue. 

We have rigorously evaluated the performance in extensive real-world experiments over the course of about half a year. We included gaps, stepping stones, narrow beams, and tall boxes in our tests, and demonstrated that our method outperformed the \ac{RL} baseline controller on every single terrain. Next, we evaluated the robustness on slippery and soft ground, each time outperforming two model-based controllers. 

Furthermore, we have shown that the emerging policy can track the motion of two different planners utilizing the same trotting gait. This was possible because the observed footholds seem to be mostly invariant under the choice of the optimizer. However, certain obstacles may encourage the deployed planner to produce footprint patterns that otherwise do not emerge during training. In this case, we would expect a degraded tracking performance.

In addition to our main contribution, we have demonstrated several other notable results. \highlight{First}, our policy, which was trained exclusively with visual perception, is still able to generalize to blind locomotion. \highlight{Second}, A simple \ac{MLP} trained with an asymmetric actor/critics setup achieves similar robust behaviors as much more complex teacher/student trainings~\cite{Lee2020,Miki2022}. \highlight{Third}, Our locomotion policy can handle a lot of noise and drift in the visual data without relying on complicated gaited networks, which might be difficult to tune and train~\cite{Miki2022}.

Contrary to our expectations, the proposed training environment was found to \highlight{not be} more sample efficient than similar unifying \ac{RL} approaches~\cite{Miki2022,Rudin2021}. The large number of epochs required for convergence suggests that foothold accuracy is something intrinsically complicated to learn.

\highlight{In this work, we emphasized that \ac{TO} and \ac{RL} share complementary properties and that no single best method exists to address the open challenges in legged locomotion. The proposed control architecture leverages this observation by combining the planning capabilities of the former and the robustness properties of the latter. It does, by no means, constitute a universal recipe to integrate the two approaches in an optimal way for a generic problem. Moreover, one could even extend the discussion with self- and unsupervised learning, indirect optimal control, dynamic programming, and stochastic optimal control. Nevertheless, our results may motivate future research to incorporate the aspect of planning into the concept \ac{RL}.}

We see several promising avenues for future research. Many successful data-driven controllers have the ability to alter the stride duration of the trotting gait. We expect a further increase in survival rate and technical skills if the network policy could suggest an arbitrary contact schedule to the motion optimizer. Moreover, a truly hybrid method, in which the policy can directly modify the cost function of the planner, may be able to generate more diversified motions. Our results indicate that \ac{IK} is difficult to learn. To increase the sample efficiency and improve generalization across different platforms, a more sophisticated network structure could exploit prior knowledge of analytical \ac{IK}. Another potential research direction may focus on leveraging the benefits of sampling trajectories from an offline buffer. This could \highlight{substantially} reduce the training time and allow for the substitution of \ac{TAMOLS} with a more accurate \ac{TO} method, or even expert data gathered from real animals.

\section*{Materials and Methods}
\subsection*{\highlight{Motivation}}
To motivate the specific architectural design, we first identify the strengths and weaknesses of the two most commonly used control paradigms in legged locomotion.

\ac{TO} amounts to open-loop control, which produces suboptimal solutions in the presence of stochasticity, modeling errors, and small prediction windows. Unfortunately, these methods also introduce many assumptions, mostly to reduce computation time or achieve favorable numerical properties. For instance, the feet are almost always pre-selected interaction points to prevent complex collision constraints, contact and actuator dynamics are usually omitted or smoothed out to circumvent stiff optimization problems, and the contact schedule is often pre-specified to avoid the combinatorial problem imposed by the switched system dynamics. Despite a large set of strong assumptions, real-time capable planners are not always truly real-time. The reference trajectories are updated around $\unit[5]{Hz}$~\cite{Melon2021} to $\unit[100]{Hz}$~\cite{Ruben2022} and realized between $\unit[400]{Hz}$ to \unit[1000]{Hz}. In other words, these methods do not plan fast enough to catch up with the errors they are making. While structural~\cite{Kalakrishnan2010} or environmental~\cite{Griffin2019, Ruben2022} decomposition may further contribute to the overall suboptimality, they were found useful for extracting good local solutions on sparse terrains.
Because the concept of planning is not restricted to the tuning domain, model-based approaches tend to generalize well across different terrain geometries~\cite{Jenelten2022,Ruben2022}. Moreover, since numerical solvers perform very cheap and sparse operations on the elevation map, the map resolution can be arbitrarily small, facilitating accurate foothold planning.

\ac{RL}, on the other hand, leads to policies that represent global closed-loop control strategies. Deep neural networks are large capacity models, and as such, can represent locomotion policies without introducing any assumption about the terrain or the system \highlight{(except from being Markovian)}. They exhibit good interpolation in-between visited states but do not extrapolate well to unseen environments. Despite their large size, the inference time is usually relatively small. The integration of an actuator model has been demonstrated to improve sim-to-real-transfer~\cite{Hwangbo2019}, while the stochasticity in the system dynamics and training environment can effectively be utilized to synthesize robust behaviors~\cite{Lee2020,Miki2022}. Contrary to\ac{TO}, the elevation map is typically down-sampled~\cite{Miki2022,Rudin2021} to avoid immense memory consumption during training.

In summary, \ac{TO} might be better suited if good generalization and high accuracy are required, whereas \ac{RL} is the preferred method if robustness is of concern or onboard computational power is limited. As locomotion combines challenges from both of these fields, we formulate the goal of this work as follows: \ac{RL} shall be used to train a low-level tracking controller that provides \highlight{markedly} more robustness than classical inverse dynamics. The accuracy and planning capabilities of model-based \ac{TO} shall be leveraged on a low level to synthesize a unifying locomotion strategy that supports diverse and generalizing motions.

\subsection*{Reference Motions}
Designing a \ac{TO} problem for control always involves a compromise, that trades off physical accuracy and generalization against good numerical conditioning, low computation time, convexity, smoothness, availability of derivatives, and the necessity of a high-quality initial guess. In our work, we generate the trajectories using \ac{TAMOLS}~\cite{Jenelten2022}. Unlike other similar methods, it does not require terrain segmentation nor pre-computation of footholds, and its solutions are robust under varying initial guesses. The system dynamics and kinematics are simplified, allowing for fast updates. During deployment, we also compare against baseline-to-2, which builds up on more complex kinodynamics. Due to the increased computation time and in particular the computationally demanding map-processing pipeline, this method is not well-suited to be used directly within the learning process (the training time would be expected to be about eight times larger).

We added three crucial features to \ac{TAMOLS}: First, we enable parallelization on CPU, which allows multiple optimization problems to be solved simultaneously. Second, we created a python interface using \texttt{pybind11}~\cite{pybind11}, enabling it to run in a python-based environment. Finally, we assume that the measured contact state always matches the desired contact state. This renders the \ac{TO} independent of contact estimation, which typically is the most fragile module in a model-based controller.

The optimizer requires a discretized $2.5$d representation of its environment, a so-called elevation map, as input. We extract the map directly from the simulator by sampling the height across a fixed grid. For both training and deployment, we use a fixed trotting gait with a stride duration of $\unit[0.93]{s}$ and swing phase of $\unit[0.465]{s}$, and set the resolution of the grid map to $\unit[0.04\times 0.04]{m^2}$.

\subsection*{Overview of the Training Environment}
The locomotion policy $\pi(\vec a\mid \vec o)$ is a stochastic distribution of actions $\vec a \in \mathcal{A}$ that are conditioned on observations $\vec o \in \mathcal{O}$, parametrized by an \ac{MLP}. The action space comprises target joint positions that are tracked using a PD controller, following the approach in~\cite{Hwangbo2019} and related works~\cite{Lee2020,Miki2022,Rudin2022}. 

\begin{figure*}
\centering
\includegraphics[width=7.3in]{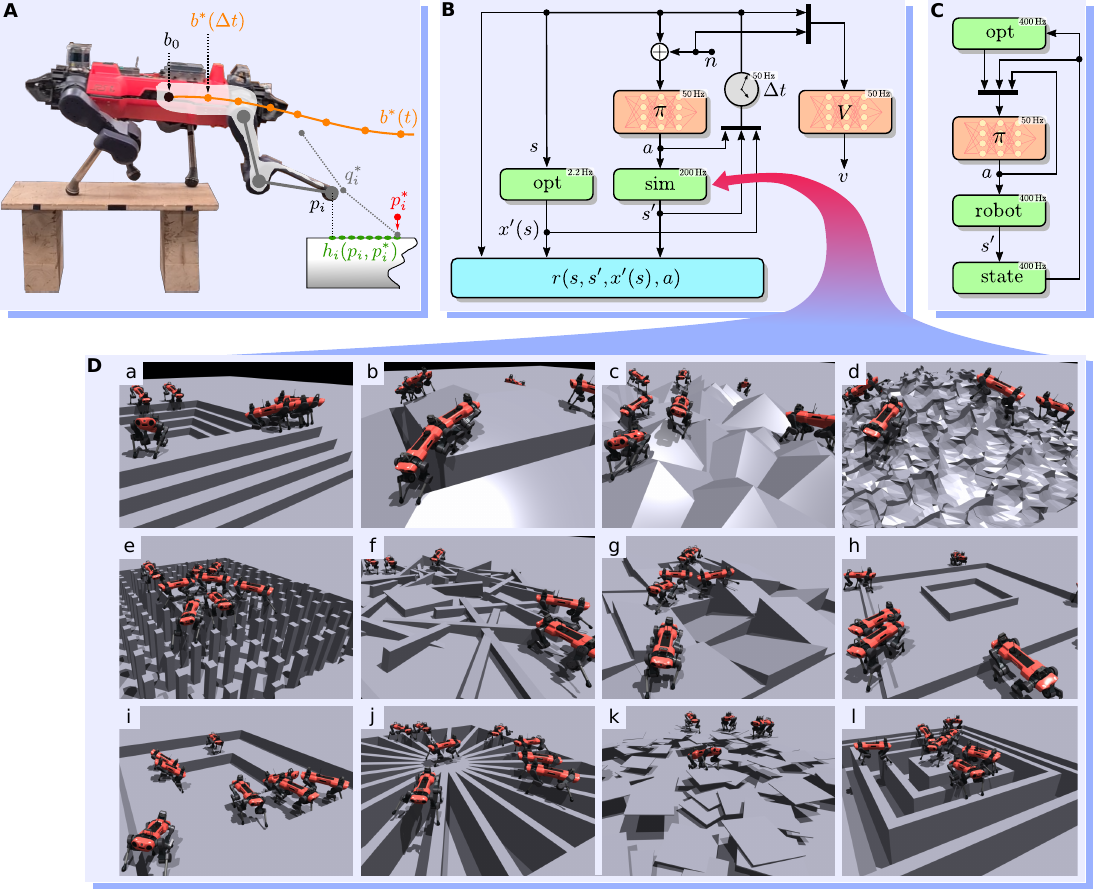}
\caption{\textbf{\highlight{Overview of the training method and deployment strategy.}} \textbf{(A)} The optimized solution provides footholds $\vec p_i^*$, desired base pose $\vec b^*$, twist $\dot {\vec b}^*$, and acceleration $\ddot {\vec b}^*$ (extracted one policy step $\Delta t$ ahead), as well as desired joint positions $\vec q^*$. Additionally, a height scan $h$ is sampled between the foot position $\vec p_i$ and the corresponding foothold.
\textbf{(B)} Training environment: The optimizer runs in parallel to the simulation. At each leg touch-down, a new solution $\vec x'$ is generated. The policy $\pi$ drives the system response $\vec s'$ toward the optimized solution $\vec x'(\vec s)$, which is encouraged using the reward function $r$. Actor observations are perturbed with the noise vector $\vec n$, while critics and the \ac{TO} receive ground truth data.
\textbf{(C)} Deployment: Given the optimized footholds, the network computes target joint positions that are tracked using a PD control law. The state estimator (state) returns the estimated robot state, which is fed back into the policy and the optimizer. \textbf{(D)} The list of terrain types includes a) stairs, b) combinations of slopes and gaps, c) pyramids, d) slopped rough terrain, e) stepping stones, f) objects with randomized poses, g) boxes with tilted surfaces, h) rings, i) pits, j) beams, k) hovering objects with randomized poses, and l) pallets.}
\label{fig:intro:title_image}
\end{figure*}

Given the state $\vec s \in \mathcal{S}$, we extract the solution at the next time step $\vec x'(\vec s) \in \mathcal{X} \subseteq \mathcal{S}$ from the optimizer, which includes four footholds $\vec p^*_{i={0,\ldots, 3}}$, joint positions $\vec q^*$ at touch-down time, and the base trajectory evaluated at the next time step. The base trajectory consists of of base pose $\vec b^*(\Delta t)$, twist $\dot{\vec b}^*(\Delta t)$, and linear and angular acceleration $\ddot{\vec b}^*(\Delta t)$. More details can be found in Fig.~\ref{fig:intro:title_image} A. We then sample an action from the policy. It is used to forward simulate the system dynamics, yielding a new state $\vec s'\in \mathcal{S}$, as illustrated in Fig.~\ref{fig:intro:title_image}~B.

To define a scalar reward $r(\vec s,\vec s',\vec x',\vec a)$, we use a monotonically decreasing function of the error between the optimized and measured states, \highlight{that is} $r \propto \vec x'(\vec s) \ominus\vec x(\vec s')$. The minus operator $\ominus$ is defined on the set $\mathcal{X}$, the vector $\vec x'(\vec s)$ is the optimized state, and $\vec x(\vec s')$ is the state of the simulator after extracting it on the corresponding subset. The policy network can also be understood as a learned model reference adaptive controller with the optimizer being the reference model.

In this work, we use an asymmetrical actor/critic method for training. The value function approximation $V(\vec o,\vec {\tilde o})$ uses privileged $\vec {\tilde o} \in \tilde {\mathcal{O}}$ as well as policy observations $\vec o$.

\subsection*{Observation Space}
The value function is trained on policy observations and privileged observations, while the policy network is trained on the former only~\cite{Brakel2021}. All observations are given in the robot-centric base frame. The definition of the observation vector is given below, \highlight{whereas} noise distributions and dimensionalities of the observation vectors can be found in \highlight{the supplementary methods and Table~\ref{tab:observations}}.

\subsubsection*{Policy Observations}
The policy observations comprise proprioceptive measurements such as base twist, gravity vector, joint positions, and joint velocities. The history only includes previous actions~\cite{Rudin2021}. Additional observations are extracted from the model-based planner, including planar coordinates of foothold positions ($xy$ coordinates), desired joint positions at touch-down time, desired contact state, and time left in the current phase. The latter two are per-leg quantities that fully describe the gait pattern. Footholds only contain planner coordinates since the height can be extracted from the height scan.

The height scan, which is an additional part of the observation space, enables the network to anticipate a collision-free swing leg trajectory. In contrast to similar works, we do not construct a sparse elevation map around the base~\cite{Tsounis2020,Rudin2021} or the feet~\cite{Miki2022}. Instead, we sample along a line connecting the current foot position with the desired foothold (Fig.~\ref{fig:intro:title_image} A). This approach has several advantages: \highlight{First}, the samples can be denser by only scanning terrain patches that are most relevant for the swing leg. \highlight{Second}, it prevents the network from extracting other information from the map, which is typically exposed to most uncertainty (for instance, occlusion, reflection, odometry drift, discretization error, etc.). \highlight{Third}, it allows us to conveniently model elevation map drift as a per-foot quantity, \highlight{which means that} each leg can have its own drift value.

We use analytical \ac{IK} to compute the desired joint positions. As the motion optimizer may not provide a swing trajectory, as is the case for \ac{TAMOLS}, we completely skip the swing phase. This means that the \ac{IK} is computed with the desired base pose and the measured foot position for a stance leg, and the target foothold for a swing leg. 

It is worth noting that we do not provide the base pose reference as observation. This was found to reduce sensitivity to mapping errors and to render the policy independent of the utilized planner.
Finally, to allow the network to infer the desired walking direction, we add the reference twist (before optimization) to the observation space.

\subsubsection*{Privileged Observations}
The privileged observations contain the optimized base pose, base twist, and base linear and angular acceleration, extracted one time step ahead. In addition, the critics can observe signals confined to the simulator, such as the external base wrench, external foot forces, the measured contact forces, friction coefficients, and elevation map drift.

\subsection*{Reward Functions}
The total reward is computed as a weighted combination of several individual components, which can be categorized as follows: ``tracking'' of reference motions, encouraging ``consistent'' behavior, and other ``regularization'' terms necessary for successful sim-to-real transfer. The reward functions are explained below whereas weights and parameters are reported in Table~\ref{tab:rewards}.

\subsubsection*{Base Pose Tracking}
To achieve tracking of the reference base pose trajectory, we use 
\begin{equation} \label{eq:r:base}
    r_{Bn} = e^{-\sigma_{Bn} \cdot ||{\vec b}^*{}(t + \Delta t)^{(n)} \ominus {\vec b}(t)^{(n)}||^2},
\end{equation}
where $n=\{0,1,2\}$ is the derivative order, $\vec b(t)$ is the measured base pose, $\vec b^*(t + \Delta t)$ is the desired base pose sampled from the reference trajectory one policy step $\Delta t$ ahead, and $\ominus$ denotes the quaternion difference for base orientation and the vector difference otherwise. We refer to the above reward function as a ``soft'' tracking task because large values can be scored even if the tracking error does not perfectly vanish. 

To further analyze the reward function, we decompose the base trajectory into three segments. The ``head'' starts at time zero, the ``tail'' stops at the prediction horizon, and the ``middle'' connects these two segments with each other.  
A logarithmic reward function would prioritize the tracking of the trajectory head, while a linear penalty would focus on making progress along the whole trajectory at once. Contrary, the exponential shape of the reward function splits the tracking task into several steps. During the initial epochs, the tracking error of the trajectory middle and tail will likely be relatively large, and thus, do not contribute \highlight{notably} to the reward gradient. As a result, the network will minimize the tracking error of the trajectory head. Once its \highlight{effect} on the gradient diminishes, the errors corresponding to the trajectory middle will dominate the gradient landscape. In the final training stages, tracking is mostly improved around the trajectory tail.

\subsubsection*{Foothold Tracking}
We choose a logarithmic function
\begin{equation} \label{eq:r:footholds}
    r_{pi} = -\ln(||\vec p_i^* - \vec p_i||^2 + \epsilon),
\end{equation}
to learn foothold tracking, where $\vec p_i$ is the current foot position of leg $i\in\{0,\ldots,3\}$, $\vec p_i^*$ is the corresponding desired foothold, and $ 0 < \epsilon \ll 1$ is small number ensuring the function is well defined. The above reward function may be termed a ``hard'' tracking task, as the maximum value can only be scored if the error reaches zero. As the tracking improves, the gradients will become larger, resulting in even tighter tracking toward the later training stages. 

A dense reward structure typically encourages a stance foot to be dragged along the ground to further minimize the tracking error. To prevent such drag motions from emerging, the above reward is given for each foot at most once during one complete gait cycle: more specifically, if and only if the leg is intended to be in contact and the norm of the contact force indicates a contact, \highlight{that is} if $||\vec f_i|| > 1$, then the agent receives the reward. 

\subsubsection*{Consistency}
In \ac{RL} for legged locomotion, hesitating to move over challenging terrains is a commonly observed phenomenon that prevents informative samples from being gathered and thus impedes the agent's performance. This behavior can be explained by insufficient exploration: The majority of agents fail to solve a task while a small number of agents achieve higher average rewards by refusing to act. To overcome this local optimum, we propose to encourage consistency by rewarding actions that are intended by previous actions. In our case, we measure consistency as the similarity between two consecutive motion optimizations. If the solutions are similar, the agent is considered to be ``consistent''. We measure similarity as the Euclidean distance between two adjacent solutions and write
\begin{equation}
r_{c} = \sum_{\delta t j + t_0 \in (T_a \cap T_b)} -\delta t||\vec b_a^*(\delta t j+ t_{0,a}) \ominus \vec b_b^*(\delta t j + t_{0,b})|| - w_p ||\vec p_a^* - \vec p_b^*||.
\end{equation}
Here, $\vec p_t^*$ with $t = \{a,b\}$ is a vector of stacked footholds, $w_p > 0$ is a relative weight, $\delta t=\unit[0.01]{s}$ is the discretization time of the base trajectory, and $t_{0}$ is the time elapsed since the solution was retrieved. The index $a$ refers to the most recent solution, while $b$ refers to the previous solution. It is important to note that the two solution vectors $\vec x_a$ and $\vec x_b$, from which we extract the base and footholds, are only defined on their respective time intervals given by the optimization horizon $\tau_h$, i.e, $t_a \in T_a = [0,\tau_{h,a}]$ and $t_b\in T_b = [0,\tau_{h,b}]$.

\subsubsection*{Regularization}
To ensure that the robot walks smoothly, we employ two different penalty terms enforcing complementary constraints. The first term, $r_{r1} = -\sum_i |\vec {v}^T_i \vec f_i|$, discourages foot-scuffing and end-effector collisions by penalizing power measured at the feet. The second term, $r_{r2} = -\sum_i (\dot {\vec {q}}^T_i \vec \tau_i)^2$, penalizes joint power to prevent arbitrary motions, especially during the swing phase. Other regularization terms are stated in \highlight{the supplementary methods (section ``Implementation Details'')}.

\subsection*{Training Environment}
To train the locomotion policy, we employ a custom version of \ac{PPO}\cite{Schulman2017} and a training environment that is mostly identical to that introduced in\cite{Rudin2021}. It is explained in more detail in \highlight{the supplementary methods (section ``Training Details'') and Table~\ref{tab:ppo}}. Simulation and back-propagation are performed on GPU, while the optimization problems are solved on CPU.

\subsubsection*{Termination}
We use a simple termination condition where an episode is terminated if the base of the robot makes contact with the terrain.

\subsubsection*{Domain Randomization}
We inject noise into all observations except for those designated as privileged. At each policy step, a noise vector $\vec n$ is sampled from a uniform distribution and added to the observation vector, with the only exceptions of the desired joint positions and the height scan.

For the elevation map, we add noise before extracting the height scan. The noise is sampled from an approximate Laplace distribution where large values are less common than small ones. We perturb the height scan with a constant offset, which is sampled from another approximate Laplace distribution for each foot separately. Both perturbations discourage the network to rely extensively on perceptive feedback and help to generalize to various perceptive uncertainties caused by odometry drift, occlusion, and soft ground.

All robots are artificially pushed by adding a twist offset to the measured twist at regular time instances. Friction coefficients are randomized per leg once at initialization time. To render the motion robust against disturbances, we perturb the base with an external wrench and the feet with external forces. The latter slightly stiffens up the swing motion but improves tracking performance in the presence of unmodeled joint frictions and link inertia. The reference twist is resampled in constant time intervals and then held constant. 

The solutions for the \ac{TO} problems are obtained using ground truth data, which include the true friction coefficients, the true external base wrench, and noise-free height map. In the presence of simulated noise, drift, and external disturbances, the policy network is therefore trained to reconstruct a base trajectory that the optimizer would produce given the ground truth data. However, there is a risk that the network learns to remove the drift from the height scan by analyzing the desired joint positions. During hardware deployment, such a reconstruction will fail because the optimizer is subject to the same height drift. To mitigate this issue, we introduce noise to the desired joint position observations, sampled from a uniform distribution with boundaries proportional to the drift value. 

\subsubsection*{Terrain Curriculum}
We use a terrain curriculum as introduced in~\cite{Rudin2021}. Before the training process, terrain patches of varying types and difficulties are generated, and each agent is assigned a terrain patch. As an agent acquires more skills and can navigate the current terrain, its level is upgraded, \highlight{which means} it will be re-spawned on the same terrain type, but with a harder difficulty. We have observed that the variety of terrains encountered during training influences the sim-to-real transfer. We thus have included a total of $12$ different terrain types with configurable parameters (Fig.~\ref{fig:intro:title_image}~D), leading to a total of $120$ distinguishable terrain patches. The terrain types classify different locomotion behaviors, s.a. climbing (``stairs'', ``pits'', ``boxes'', ``pyramids''), reflexing (``rough'', ``rings'', ``flying objects''), and walking with large steps (``gaps'', ``pallets'', ``stepping stones'', ``beams'', ``objects with randomized poses''). Our terrain curriculum consists of $10$ levels, where one of the configurable parameters is modulated to increase or decrease its difficulty. This results in a total of $1200$ terrain patches, each with a size of $\unit[8\times8]{m^2}$, summing up to a total area of $\unit[76800]{m^2}$, which is approximately the size of $14$ football fields or $10$ soccer fields.

\subsection*{Training}
Solving the \ac{TO} problems at the policy frequency during training was found to provoke poor local optima. In such a case, the optimizer adapts the solution after each policy step: If the agent is not able to follow the reference trajectory, the optimizer will adapt to the new state s.t. the tracking problem becomes feasible again. This means that the agent can exhibit ``lazy'' behavior and still collect some rewards. We prevent such a local optimum by updating the optimizer only at a leg touch-down (after $0.465$ seconds). This also greatly reduces learning time because computational costs are reduced by a factor of $23$.
After a robot fell (on average, once every $18$ seconds), was pushed (after $10$ seconds) or its twist commands changed (three times per episode), the optimized trajectories are no longer valid. To guarantee that the locomotion policy generalizes across different update rates, we additionally recompute the solution in all those scenarios. 

We trained the policy with a massive parallelization of $64^2=4096$ robots, for a total of $90000$ epochs. Each epoch consisted of $45$ learning iterations where each iteration covered a duration of $0.02$ seconds. Considering the variable update rate explained previously, this resulted in a total of $8295$ days (or $23$ years) of optimized trajectories. The policy can be deployed after about one day of training ($6000$ epochs), reaches $\unit[90]{\%}$ of its peak performance after three days ($20000$ epochs), and is fully converged after two weeks ($90000$ epochs).

In comparison, the baseline-rl-1 policy was trained for $4000$ epochs with $1000$ parallelized robots over $5$ consecutive days. Each epoch lasted for $5$ seconds, resulting in a throughput of $46$ simulated seconds per second. Our policy was trained for $14$ days, with each epoch lasting for $0.9$ seconds, leading to a throughput of $27$ simulated seconds per second. Thus, despite generating $1.6$ years of desired motions per day, our approach has only a $1.7$ times lower throughput than the baseline.

\subsection*{Deployment}
We deploy the policy at a frequency of $\unit[50]{Hz}$ without any fine-tuning. The motion optimizer runs at the largest possible rate in a separate thread. For \ac{TAMOLS} with a trotting gait, this is around $\unit[400]{Hz}$ and for baseline-to-2 around $\unit[100]{Hz}$ (both are faster than the policy frequency). At each step, the policy queries the most recent solution from the thread pool and extracts it $\Delta t = \unit[0.02]{s}$ ahead of the most recent time index. 

For our experiments, we used three different types of ANYmal robots~\cite{Hutter2016}, two version C and one version D, for which we trained different policies. ANYmal C is by default equipped with four Intel RealSense D435 depth cameras whereas ANYmal D has eight depth cameras of the same type. For the second Version C, the depth cameras were replaced with two identical Robosense Bpearl dome LiDAR sensors. \highlight{For the outdoor experiments, we mostly used this robot, as the Bpearls tend to be more robust against lighting conditions.} Motion optimization and the forward propagation of the network policy are done on a single Intel core-i7 8850H machine. Elevation mapping~\cite{elevationmapping_cupy} runs on a dedicated onboard Nvidia Jetson.



\section*{Acknowledgments}
\textbf{Funding:}
This project has received funding from the European Research Council (ERC) under the European Union’s Horizon 2020 research and innovation programme grant agreement No 852044. This research was supported by the Swiss National Science Foundation (SNSF) as part of project No 188596, and by the Swiss National Science Foundation through the National Centre of Competence in Research Robotics (NCCR Robotics). \textbf{Author Contribution:} 
F.J. formulated the main ideas, trained and tested the policy using baseline-to-1, and conducted most of the experiments. J.H. interfaced baseline-to-2 with the tracking policy and conducted the box-climbing experiments. F.F. contributed to the Theory and improved some of the original ideas. All authors helped to write, improve, and refine the paper. \textbf{Competing interests:} The authors declare that they have no competing interests. \textbf{Data and materials availability:} All (other) data needed to evaluate the conclusions in the paper are present in the paper or the Supplementary Materials. \highlight{Data-sets and code to generate all our figures are made publicly available~\cite{DatasetJenelten}.}

\clearpage

\section*{\highlight{Supplementary Methods}}
\subsection*{Nomenclature}
$(\cdot)_i$ index for leg $i\in\{0,\ldots,3\}$\\
$\tau_h$ prediction horizon\\
$\vec b^*(t)$ optimized base pose trajectory\\
$\dot {\vec b}^*(t)$ optimized base twist trajectory\\
$\ddot {\vec b}^*(t)$ optimized base acceleration trajectory\\
$\vec q^*$ optimized joint positions obtained from \acs{IK}\\
$\vec p_i^*$ optimized foothold\\
$\vec p_t$ vector of stacked footholds, computed at time $t$\\
$\vec p_i$ measured foot position\\
$\vec f_i$ measured contact force\\
$\vec v_i$ measured foot velocity\\
$\vec q_i$ measured joint positions\\
$\vec \tau_i$ measured joint torques\\
$\vec s$ state of the robot \\
$\vec x'(\vec s)$ solution of optimization problem one time-step ahead, initialized with state $\vec s$\\
$\vec x(\vec s')$ state $\vec s'$ extracted on subset of the optimizer\\
$\vec o$ policy observations\\
$\vec {\tilde o}$ privileged observations\\
$\vec a$ actions

\subsection*{Training Details}
We use a simulation time step of $5$ ms and a policy time step of $20$ ms.

The elevation map noise is sampled from a Laplace distribution, which is approximated by two uniform distributions. We first sample the boundaries of the uniform distribution $h_{n,max} \sim \unit[\mathcal{U}(0, 0.2)]{m}$, and then sample the noise from $h_{n} \sim \mathcal{U}(-h_{n,max}, h_{n,max})$.
Similar holds for the height drift: We first sample the boundaries $h_{d,max} \sim \unit[\mathcal{U}(0, 0.2)]{m}$ and then sample five drift values $h_{d0},\ldots,h_{d4} \sim \mathcal{U}(-h_{d,max}, h_{d,max})$, where $h_{d0}$ is constant for the entire map and $h_{d1}.\ldots,h_{d4}$ is sampled for each foot individually. The total drift $h_d$ is obtained by summing up the map-drift and the per-foot drift. Elevation map noise and drift are re-sampled after a constant time interval of $\unit[8]{s}$.

The noise vector added to the desired joint positions depends on the height drift, and is sampled from the uniform distribution $q_{n} \sim 2 \cdot \mathcal{U}(-h_{d}, h_{d})$.

We push the robots after $\unit[10]{s}$ by adding a twist offset to the measured twist sampled from $\Delta \dot {\vec b} \sim \mathcal{U}\unit{(-1, 1)}[m/s]$. Friction coefficients are randomized per leg and sampled from $\mu \sim \mathcal{U}(0.1, 1.2)$. For each episode, we sample an external base wrench $\vec {\tau}_B \sim \mathcal{U}(-15, 15)$ and a external foot force $\vec f_{ee} \sim \unit[\mathcal{U}(-2, 2)]{N}$.

The reference twist is re-sampled from a uniform distribution three times per episode. A third of the robots has zero lateral velocity \highlight{whereas} the heading velocity is sampled from $v_x \sim \unit[\mathcal{U}(-1, 1)]{m/s}$. Another third has zero heading velocity and its lateral velocity is sampled from $v_y \sim \unit[\mathcal{U}(-0.8, 0.8)]{m/s}$. The last third has mixed heading and lateral velocities sampled from their respective distributions. For all three cases, the yaw velocity is sampled from $v_\psi \sim \unit[\mathcal{U}(-0.8, 0.8)]{rad/s}$.

The PPO hyperparameters used for the training are stated in Table~\ref{tab:ppo}. All policies were trained with a seed of $1$.
\begin{table}[]
\centering
\begin{tabular}{r|l}
parameter type & number \\ \hline 
batch size & $45 \cdot 4096 = 184320$ \\
mini batch size & $4 \cdot 4096 = 16384$ \\
number of epochs & $5$ \\
clip range & $0.2$ \\
entropy coefficient & $0.0035$ \\
discount factor & $0.99$ \\
GAE discount factor & $0.95$ \\
desired KL-divergence & $0.01$ \\
learning rate & adaptive \\
\end{tabular}
\caption{\textbf{PPO hyperparameters}. The value $4096$ is the number of parallelized environments. All policies were trained with the same parameters.}
\label{tab:ppo}
\end{table}

\subsection*{Implementation Details}
To parameterize the policy (or actor) network, a Gaussian distribution is used, where the mean is generated by an MLP parametrized by $\theta$. The standard deviation is added independently of the observations as an additional layer in the network. Specifically, the policy can be written as $\pi(\vec a\mid \vec o) \sim \mathcal{N}(\vec \mu_\theta(\vec o), \vec \sigma_\theta)$, where $\vec \mu_\theta(\vec o)$ represents the mean and $\vec \sigma_\theta$ the standard deviation. The value function (or critics) is generated by another MLP as  $V(\vec o, \vec{\tilde o}) \sim \mathcal{N}(\vec \mu_\phi(\vec o, \vec{\tilde o}), \vec 0)$, which is parametrized by $\phi$.
For both MLP's we use three hidden layers with each having $512$ neurons. Observations are normalized using running means and running standard deviation.
\begin{table}[]
\centering
\begin{tabular}{l|l|l|l}
type & observations & dim & noise \\ \hline
policy & base twist & 6 & \\
& gravity vector & 3 & $\pm0.05$ \\
& joint positions & 12 & $\pm0.01$ \\
& joint velocities & 12 & $\pm 1.5$ \\
& previous actions & 12 & $0$ \\
& planar footholds & 8 &$\pm0.05$ \\
& desired joint positions & 12 & Laplace  \\
& desired contact state & 4 & \\
& time left in phase & 4 & \\
& reference twist & 3 & \\
& height scan & 40 & Laplace \\ \hline
privileged & desired base position & 3& \\
& desired base quaternion & 4& \\
& desired base twist & 12& \\
& consistency reward & 1& \\
& external base wrench & 6& \\
& external foot force & 12& \\
& friction coefficients & 4& \\
& height drift & 4&
\end{tabular}
\caption{\textbf{Policy and privileged observations}. The column ``noise'' contains the upper and lower values of the uniform distribution, with units identical to that one of the observations. Entries marked with ``Laplace'' indicate the noise distribution is sampled from an approximate Laplace distribution as explained in \highlight{the supplementary methods (section ``Training Details'')}. A missing value indicates that the noise level is zero.}
\label{tab:observations}
\end{table}

Policy and privileged observations can be found in Table~\ref{tab:observations}, and relative weights of the reward functions are given in Table~\ref{tab:rewards}. The implementation of most of the reward functions are given in the main text. In addition to those, we penalize the action rate $r_a = -||\vec a_t - \vec a_{t-1}||^2$ and joint acceleration $r_{\ddot {\vec q}} = -||\ddot {\vec q}||^2$, with $\vec a_t, \vec a_{t-1}$ being the current and previous actions, and $\ddot {\vec q}$ the measured joint accelerations.
\begin{table}[]
\centering
\begin{tabular}{l|l|l|l}
reward type & reward name & weight & parameter \\ \hline
tracking    & base position & $1$ & $\sigma = 1200$ \\
            & base rotation & $1$ & $\sigma = 90$ \\
            & base linear velocity & $1$ & $\sigma = 10$ \\
            & base angular velocity & $1$ & $\sigma = 1$ \\
            & base linear acceleration & $1$ & $\sigma = 0.05$ \\
            & base angular acceleration & $1$ & $\sigma = 0.005$ \\ 
            & footholds & $6$ & $\epsilon=10^{-5}$ \\ \hline
consistent behavior & consistency & $20$ & \\ \hline
regularization & foot power & $-0.02$ & \\
               & joint power & $-0.025$ & \\
               & action rate & $-0.02$ & \\
               & joint acceleration & $-10^{-6}$ 
\end{tabular}
\caption{\textbf{Reward functions}. }
\label{tab:rewards}
\end{table}

\subsection*{Experimental Setup for Evaluation of Optimizer Rate}
The experimental setup mostly coincided with the training environment \highlight{as detailed in the supplementary methods (section ``Training Details'')}, with four minor differences: The terrain curriculum was disabled, and for each terrain type, we selected the most difficult one. A push or a velocity change did not trigger a new solution to be computed. The optimizer was not informed about the true values for friction coefficients and external base wrenches. Instead, nominal values were used. All robots were walking in heading direction with a reference velocity sampled uniformly from the interval $\unit[0.8 \ldots 0.95]{m/s}$. The robots were pushed after $\unit[5]{s}$ with a disturbance twist sampled from the interval $\Delta \dot {\vec b} \sim \mathcal{U}\unit[(-2, 2)]{m/s}$, \highlight{which is} twice as much as experienced during training. We did not add external foot forces, noise, or map drift. For each optimizer rate in $\unit[\{1, 2, 2.5, 3.3, 5, 10, 20, 50\}]{Hz}$, the success and failure rates were averaged over $8$ different experiments conducted with varying seeds selected from the interval $\{2,\ldots,9\}$. Each experiment was conducted with $4096$ robots, distributed across $120$ terrains, lasting for one episode ($=\unit[20]{s}$). 

\subsection*{Experimental Setup for Performance Evaluation}
Baseline-rl-3 used the same observations and policy structure as baseline-rl-2. But it was trained within the same environment as ours using a comparable number of epochs.

For the evaluation on all terrains, we used the training environment \highlight{as detailed in the supplementary methods (section ``Training Details'')}, with the following simplifications: The terrain curriculum was disabled, and for each terrain type, we selected the most difficult one. All robots were walking in heading direction with a reference velocity sampled uniformly from the interval $\unit[0.8 \ldots 0.95]{m/s}$. We did not add external foot forces, noise, or map drift. The friction coefficients between feet and ground were set to $1$. In addition, for the evaluation on specific terrains, we did not push the robots, and we did not re-sample the reference velocities. This setup relaxes the locomotion task on topologically hard terrains such as stepping stones, that otherwise might not be traversable. Results were averaged over eight different seeds.

\subsection*{Experimental Setup for Performance Evaluation under Drift}
The experimental setup mostly coincided with the training environment \highlight{as detailed in the supplementary methods (section ``Training Details'')}, with the following differences: The terrain curriculum was disabled, and for each terrain type, we selected the most difficult one. All robots were walking in heading direction with a reference velocity sampled uniformly from the interval $\unit[0.8 \ldots 0.95]{m/s}$. We did not add external foot forces or noise. The friction coefficients between feet and ground were set to $1$. We did not push the robots and we did not re-sample the reference velocities. In addition, for the flat ground experiment, we replaced all terrain types with a single horizontal plane. Each experiment was repeated $21$ times with a linearly increasing drift value selected from $\unit[\{0.0,0.025,\ldots, 0.5\}]{m}$. Results were averaged over eight different seeds.

\end{document}